\documentclass[12pt,a4paper]{article}
\usepackage{algorithm}
\usepackage{algorithmic}
\usepackage{graphicx}
\usepackage{amsmath}
\title{\textbf{Cross-Country Skiing Gears Classification using Deep Learning}}

\author{Aliaa Rassem, Mohammed El-Beltagy and Mohamed Saleh}
\date{}
\begin{document}
\maketitle

\begin{abstract}
Human Activity Recognition has witnessed a significant progress in the last decade. Although a great deal of work in this field goes in recognizing normal human activities, few studies focused on identifying motion in sports. Recognizing human movements in different sports has high impact on understanding the different styles of humans in the play and on improving their performance. As deep learning models proved to have good results in many classification problems, this paper will utilize deep learning to classify cross-country skiing movements, known as gears, collected using a 3D accelerometer. It will also provide a comparison between different deep learning models such as convolutional and recurrent neural networks versus standard multi-layer perceptron. Results show that deep learning is more effective and has the highest classification accuracy.
\end{abstract}

%\begin{keyword} Deep learning; human activity recognition; convolutional neural network; long-short term memory; skiing.
%\end{keyword}

\section*{1. Introduction}
Human Activity Recognition (HAR) is one of the active research areas which retrieves the actions of humans for the sake of an automatic understanding of their behaviours. There is a significant progress in the field of HAR due to its potential in many domains. For example, recognizing activities can be useful for smart homes systems, pervasive and mobile computing,  surveillance-based security, physical therapy, rehabilitation, context-aware computing, robotics, and sports' improvement \cite{chen2015survey}. HAR approaches can be divided mainly to vision based and sensor based. Although many studies investigated the vision based approach where input signals are acquired by video and depth cameras, this approach suffers from some limitations. For example, it requires high computationally cost image processing algorithms for classifying the video sequences or the visual data. On the other side, sensors are cost effective,  work in unrestricted environments and raise less privacy concerns compared to video cameras. This leads recent studies in this field to focus on the sensor based approach where movements are acquired by inertial sensors such as accelerometers or gyroscopes  to develop a multi-variate time series classification problem. 

Activity recognition applications can be utilized for classifying either daily normal activities or  sports motion. The recognition of sports motion didn't get as much attention in research as normal human activities (like walking,sitting  and driving cars) although its high impact on understanding the different styles of athletes and on improving their performance. It also offers a feedback on performance for amateur and professional athletes and helps in predicting next moves during the sport. In this paper, we focus on classifying the human motion in cross-country skiing (XCS). XCS has different skiing techniques, primarily divided into classical and skating techniques. Each technique has different motion patterns known as gears. The challenge in XCS classification, apart from walking and running activities, is the varying intensity of the same skiing gear which can be fast or slow. In this case, frequently used features like intensity and frequency are not significant in XCS as used before for normal activities classification. \cite{holst2013classification}. \par

In most of HAR studies, the used features are selected by hand but designing hand-crafted features requires domain knowledge and may cause information loss after the features selection. In recent years, deep learning (DL) has delivered a solution to the features selection problem and became the new trend in machine learning. Deep networks such as convolutional and recurrent neural networks have been successfully applied in diverse applications such as speech recognition, natural language processing, audio processing, computer vision and robotics. The main reasons behind the success of deep networks are the stacked layers in their architecture unlike traditional networks which include three layers at most. Also, these networks apply complex non-linear functions through its layers to automatically learn  the hierarchical representations of features without any domain knowledge.  Although the remarkable results that deep learning was able to achieve in many applications, it has not been fully exploited in HAR \cite{zeng2014convolutional},  \cite{yang2015deep} and \cite{chen2016lstm}.

This paper applies deep learning approaches for human motion classification for the first time in cross country skiing using sensor data. Two deep learning approaches, convolutional and recurrent networks, are applied using different network architectures under each approach. The performance of all approaches are tested versus a traditional multi layer perceptron with two layers at most. Many experiments are carried using a large dataset of skiing data collected with a 3D sensor for different skiers and approaches are compared with respect to the testing classification error.
%%%%%%%%%%%%%%%%%%%%%%%%%%%%%%%%%%%%%%%%%%%%%%%%%%%%%%%%%%%%%%%%%%%%%%%%%%%%%%%%%%%%%%%%%%%%%%%%%%%%%%%%%%%%%%%%%%%%%%%%%%%%%%%%%%%%%%%%%%%%%%%%%%%%%%%%%%%
\fontsize{11}{10}
\section*{2. Related Works and Background} \label{sec:RWAB}

This section reviews some of the sensor based activity recognition systems using deep learning, mostly the deep convolutional and recurrent networks. The recognition systems that identify daily normal activities will be shown first, then the systems that identify sports motion. Finally, few studies which considered cross country skiing by the traditional machine learning models will be described.

% CNN
Convolutional neural networks (CNNs) have the power to capture local dependencies and keep features scale invariant during the representation learning process. These two key advantages of CNN fit the characteristics of signals of normal activities in any recognition system. An activity signal always have a high correlation between neighbouring acceleration values. Moreover, any two signals with different motion intensities may represent the same activity but for two different persons \cite{zeng2014convolutional}. Therefore, CNN was extensively applied in HAR systems to achieve better recognition accuracy.  For example, a deep CNN was deployed for activity recognition in \cite{jiang2015human} and was tested versus support vector machine (SVM) and a feature selection method. The raw sensor signals were collected from accelerometers and gyroscopes and then permuted and stacked in a signal image. The CNN architecture learned low and high- level features from the constructed activity image. Results on 3 datasets showed the largest recognition accuracy values to CNN. Another HAR system in \cite{zeng2014convolutional} introduced a new modified weight sharing technique, called partial weight sharing, to improve the classification accuracy of the convolutional network. The new technique allows sharing weights to only local filters that are close to each other and aggregated together in the max-pooling layer. The modified CNN outperformed the standard CNN in many benchmark datasets using different regularization settings of weight decay, momentum and drop-out. Furthermore, in \cite{yang2015deep}, a deep CNN was built from  convolution , rectified linear unit (ReLU) , max pooling  and  normalization layers. It was tested using benchmark Opportunity Activity Recognition  and Hand Gesture datasets. According to the values of accuracy and average F-measure, the convolutional network was more effective than four other methods: SVM, K-nearest neighbour (KNN), Means and Variance and deep belief network (DBN).  Other Comparisons between CNN and standard ML techniques were conducted in \cite{gjoreskicomparing} and  \cite{ronao2016human}. In the former study, the CNN was tested versus J48, random forest (RF), SVM, KNN, and Naïve Bayes using wrist accelerometer. It achieved higher accuracy compared to RF for all of the subjects for the left wrist. However opposite results obtained when the right wrist was analyzed. In the other study, the CNN had the best accuracy values of 94.79\% using raw sensor data and 95.75\% using additional information of Fast Fourier Transform (FFT) from the data set .

%LSTM
The deep recurrent neural network (RNN) and especially the recurrent network based on long short term memory (LSTM) has also proved its effectiveness in many recognition tasks. This architecture is able to exploit the temporal dependencies in the time series data of humans' normal activities acquired by sensors \cite{hammerla2016deep}. For example, in \cite{chen2016lstm} a LSTM was applied on WISDM activity dataset and yielded final accuracy of 95\%. In \cite{lee2012mobile},  a proposed hierarchical bidirectional LSTM (BLSTM) was tested versus a standard BLSTM for gesture recognition with smartphones . This bi-directional architecture provides two paths, forwards and backwards, for each input sequence and this feeds the network with past and future context of every value in this sequence. The first level of BLSTM was responsible for classifying input  sequences to gestures and non-gestures and then, the second level classified valid sequences to final gestures labels. The hierarchical BLSTM outperformed the standard architecture with respect to the classification accuracy. In addition, a deep learning framework based on convolutional and LSTM recurrent layers was proposed in \cite{ordonez2016deep}. The framework was able to both learn feature representations and model the temporal dependencies between their activations. Experiments on opportunity and skoda activity datasets proved the superiority of the proposed framework over a baseline CNN with respect to F1 score. Another framework that integrated many deep approaches was in \cite{hammerla2016deep}. A deep neural network, a convolutional network and two flavours of LSTM recurrent networks , the forward and bidirectional were implemented. Also, a new regularization for RNN named breaks was introduced in this study. Experiments on benchmark datasets showed that RNN outperformed CNN on activities that are short in duration but have a natural ordering where the opposite happened for prolonged and repetitive activities.

% Sports
On the other hand, few systems were proposed for sports motion recognition and the majority of these systems were based on traditional machine learning algorithms as in \cite{fuji2011development} ,\cite{barshan2014recognizing},\cite{lu2016multi} and \cite{wang2015sport}. For example, in ball games, authors in \cite{fuji2011development} used gaussian fitting and regression analysis for extracting movement characteristics of tennis players. However, in \cite{wang2015sport}, a SVM based model was used for recognizing sports with single-handed swings like tennis, badminton, and ping pong. SVM was able to classify the extracted features from triaxial accelerometers after using N-median Filtering. Other studeies applied different algorithms for classifying both daily normal activities and sports motion like in \cite{barshan2014recognizing}. In this study, a comparison between different classifiers like SVM and NN was introduced for classifying activities signals from inertial and magnetic sensors. The best recognition accuracies were from NN and SVM classifiers. Also in  \cite{lu2016multi}, a multi-classifier combination model based on Quadratic SVM (QSVM) and AdaBoost Tree ensemble learning algorithm was used for daily and sports activities recognition.

For cross country skiing, \cite{holst2013classification}, \cite{ristnercomparison} and \cite{stoggl2014automatic} applied machine learning for gears classification. In \cite{holst2013classification}, a Markov chain of multivariate Gaussian distributions was used for classification, anomalies detection and clustering periodic movement patterns.  Experiments on skiing data from 14 skiers during a training race on roller skies showed the ability of the algorithm to reach an overall performance on the test set of 98\%.  Another Markov model and a k-nearest neighbour algorithm were implemented to classify classical style and free style simultaneously in \cite{ristnercomparison}. In this study, skiing data was first preprocessed by re-sampling the sensor data to have all skiing cycles with same fixed timestamps before classification. Results showed that most of the gears were correctly classified using both algorithms but according to the error rates, KNN algorithm is preferable. Finally in  \cite{stoggl2014automatic}, the gaussian filters were used with a Markov-chain machine learning procedure to identify the cross country ski-skating gear.

\section*{2.1. Long short-term memory (LSTM) }  
\begin{figure}[h!]
\centering
\includegraphics[width=3.5in]{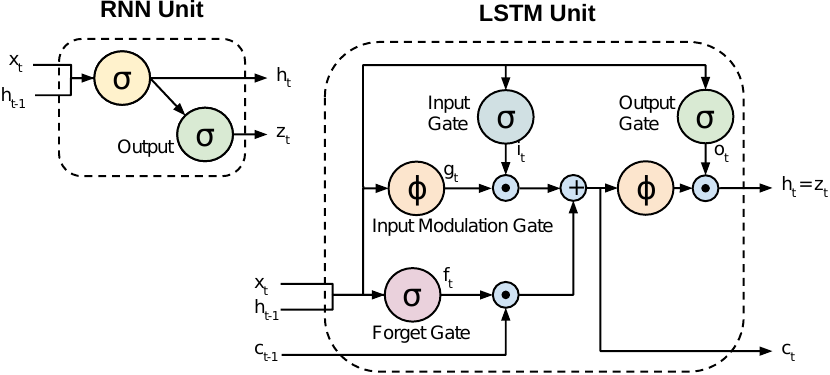}
\caption{RNN simple cell versus LSTM cell \cite{donahue2015long}}
\label{rnn}
\end{figure}

Recurrent neural networks (RNNs), figure \ref{rnn}, contain hidden layers of recurrently connected neurons to represent the temporal dependencies in data sequences. RNNs have been applied successfully in many applications such as speech recognition  and natural language processing. These networks have limitations of their own, for example it is difficult to train these networks on long input sequences. This is due to  the problems of vanishing and exploding gradients that  occur when errors are back-propagated across many time steps. The Long short term memory(LSTM) solved this problem by integrating memory units to enable learning of long temporal dynamics. Another limitation of RNNs is the dependency of outputs on previous inputs and this was solved by the bidirectional architecture \cite{graves2005framewise}. 

LSTM, figure \ref{rnn}, was firstly introduced  in 1997 by Sepp  Hochreiter \& Jürgen Schmidhuber. The LSTM cell is a variant architecture of  RNN cell which incorporates memory units that enable learning complex and long-term temporal dynamics of long sequences. These units allow to learn when to forget previous hidden states and when to update hidden states given new information. This cell includes an input $i_t$, input modulation $g_t$, forget $f_t$ and output $o_t$ gates in addition to memory cell $c_t$ and the hidden unit $h_t$, as in equations \ref{eq1} to \ref{eq6} where $W$ are the weights matrices between input or hidden state to the gates and $b$ are the biases terms.  Each block in the network has many LSTM cells where gates are shared by these cells in the block. Gates are responsible about adjusting the interactions between the memory cell and its environment. For example, the input gate allows incoming signal  to alter the state of the memory cell or blocks it where the output gate controls the effect of the memory cell on the hidden state. The  forget gate on the other side, lets the cell remembers or forgets its  previous state  \cite{chen2015survey},\cite{chen2016lstm} ,\cite{gers2002learning} and \cite{donahue2015long}.

\begin{equation}\label{eq1}
i_t= \sigma(W_{xi}x_t + W_{hi}h_{t-1}+b_i)
\end{equation}
\begin{equation}\label{eq2}
f_t= \sigma(W_{xf}x_t + W_{hf}h_{t-1}+b_f)
\end{equation}
\begin{equation}\label{eq3}
o_t= \sigma(W_{xo}x_t + W_{ho}h_{t-1}+b_o)
\end{equation}
\begin{equation}\label{eq4}
g_t= \sigma(W_{xc}x_t + W_{hc}h_{t-1}+b_c)
\end{equation}
\begin{equation}\label{eq5}
c_t= f_t \odot	 c_{t-1} +  i_t \odot	g_{t}
\end{equation}
\begin{equation}\label{eq6}
h_t= o_t \odot \phi(c_t)
\end{equation}

Some variants of LSTM were developed to enhance its performance. For example, to increase the effectiveness of the LSTM cell in learning a precise timing of the outputs, peephole connections were introduced  by Gers \& Schmidhuber in \cite{gers2002learning} . The peephole connections are used from the internal cells of the LSTM to the gates in the same cell to allow communication between gates and help the gates to access the current cell state even when the output gate is closed. Another variant of the standard LSTM cell is the bidirectional LSTM (BLSTM) where the concept of the bidirectional RNN (BRNN) were first introduced by Schuster \& Paliwal . The bidirectional architectures process each training sequence forwards and backwards by two separate recurrent nets, connected to the same output layer. Therefore, BLSTM has the ability to access long-range sequences in both input directions and access their future input information from the current state  \cite{graves2005framewise}.

\subsection*{2.2. Convolutional Neural Network (CNN) }
 
\begin{figure}[h!]
\centering
\includegraphics[width=2.9in]{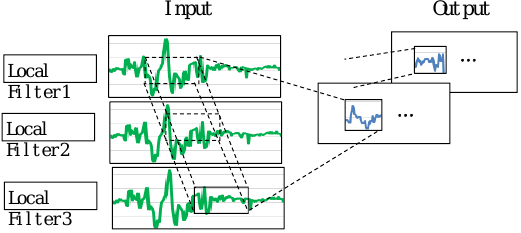}
\caption{Example of convolution filters: three feature maps at input layer and two results at output layer \cite{zeng2014convolutional}}
\label{CNN}
\end{figure}

Convolutional neural networks (CNNs), figure ~\ref{CNN} , are deep networks based on local filters which are able to discover the correlation within the input data through the convolution operation.The output feature maps from this convolution can figure out different types of features at each temporal position. Convolutional networks include mainly stacked layers of convolution and pooling operations. The convolution operation apply each local filter over all subsets of the input where weights of these filters are shared across all subsets. Then, the pooling operation splits the output features and applies some function to reduce the size of previous layer to preserve the scale invariance property of features\cite{zeng2014convolutional}. 

CNNs achieved high performance when applied in the activity recognition field. This goes back to the key advantages of CNNs, which are the local dependencies and the scale invariance. The local filters in CNN have the capability to capture local dependencies of neighbouring sensor acceleration readings of an activity signal. In addition to that, scale invariance characteristic allows CNN to successfully learn hidden features regardless of their positions or scales. This is helpful in the recognition of human activities as people may do the same activity with different speeds and intensities \cite{zeng2014convolutional}, \cite{yang2015deep}.

\section*{3. Experiments}

\begin{figure}[h!]
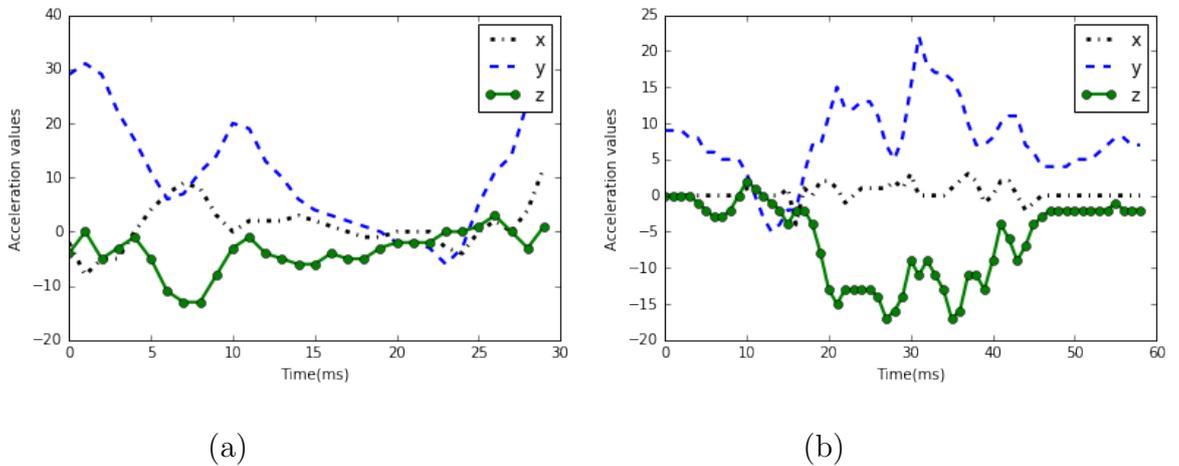

\centering
\begin{tabular}{ p{6cm} p{1cm} p{6cm} p{1cm}}
\includegraphics[width=3.0in]{figures/fig3_gear2.pdf} {\begin{center}(a)\end{center} } && \includegraphics[width=3.0in]{figures/fig3_gear3.pdf}{\begin{center}(b)\end{center} }&  \\
\end{tabular}
\caption{(a): a typical cycle of gear 2 (b): a typical cycle of gear  3}
\label{examples}
\end{figure}

All experiments were done on the cross country skiing (XCS) sport where different sessions or movement patterns are called gears. The data \footnote{racefox.se} was collected using a 3D accelerometer in form of, acceleration in the horizontal direction x, acceleration in the up direction y and in the forward direction z. The total recorded examples was 416,737 records with 50\% of records for gear 2 and 50\% for gear 3 with sampling rate 50Hz. Figure \ref{examples}, (a) and (b) shows a typical cycle of each gear in its raw format. 
%XCS has two styles, the classical and free/skate style. Each style can be divided into sub techniques, called motion patterns or gears. In this paper, we focus on skating style with its three different gears, gear 2, gear 3 and gear 4 where the used dataset includes only the first two gears. The skiing gear 2 is used for steep hills, where gear 3 used for not so steep ones and gear 4 for more flat hills. 

The data was divided into training, validation and testing sets with a segmentation process applied with window size of 1 second with 50\% overlapping. Variant models of convolutional and recurrent networks were implemented in addition to a standard MLP. For the recurrent networks, three LSTM networks architectures were applied including standard forward LSTM (LSTM-F), LSTM with peepholes (LSTM-P) and bi-directional LSTM (BLSTM) . For the convolutional networks, two CNNs were applied, the first one with only one convolution layer and the second with two convolution layers.  Finally, two multi layer perceptron (MLPs) were also tested, using one and two hidden layers. All the models were implemented using Tensorflow using different parameters settings as in table \ref{exp. results}. The maximum number of training iterations was 3000 with total number of runs was 5. The best weights associated with the minimum validation classification error were kept through the iterations with the cross-entropy error as the training error measure. The performance of models was measured by the average testing classification error over the runs where the testing error considers the number of mismatched examples.

\begin{table}
\small
\caption{ Parameters settings}\label{exp. results}
\begin{center}
\begin{tabular}{ |p{1.3cm}| p{1.5cm} p{2cm} p{1cm} p{1cm} p{1.5cm} p{1.8cm} p{1cm}|}
\hline
Model & Conv. Layers &Filters & Filter Size & Full layers & LSTM Units & Hidden Neurons & Batch Size \\
\hline
CNN & [1, 2]&[20, ,40, 50] &10 &1 & - & 1000&100\\
\hline
LSTM-F &- & - & -  & - & [25, 35, 50] &   - &100\\
\hline
LSTM-P &- & - & -  & - & [25, 35, 50] &   - &100\\
\hline
BLSTM & -& - & -  & - & [25, 35, 50] &  - &100\\
\hline
MLP& -& - & -  & [1, 2] & - & [30, 50, 100] &100\\
\hline
\end{tabular}
\end{center}
\end{table}

\section*{4. Results and Discussions}
\begin{figure}[h!]
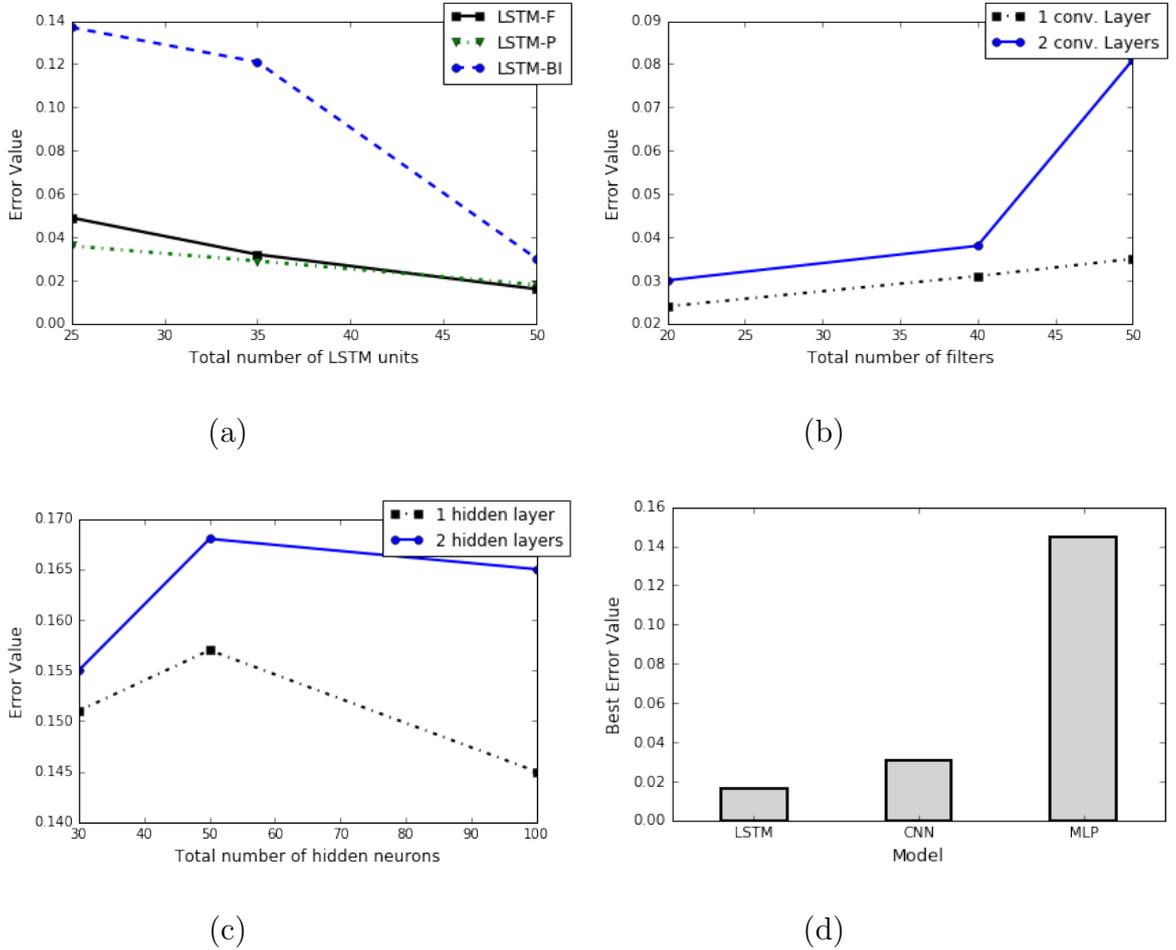

 \centering
\begin{tabular}{ p{6cm} p{1cm} p{6cm} }
\includegraphics[width=3.0in, height=2.0in]{figures/fig4_LSTM_error.pdf} {\begin{center}(a)\end{center} }& & \includegraphics[width=3.0in, height=2.0in]{figures/fig4_CNN_error.pdf} {\begin{center}(b) \end{center} }\\
\includegraphics[width=3.0in, height=2.0in]{figures/fig4_MLP_error.pdf} {\begin{center}(c) \end{center} }& &
\includegraphics[width=3.0in, height=2.0in]{figures/fig4_besterror.pdf} {\begin{center}(d) \end{center} }\\
\end{tabular}
\caption{Error Values of (a): LSTM models ,(b): CNN models, (c): MLP models, (d) Best Error}
\label{error}
\end{figure}

The testing classification error values of each model are shown in figure ~\ref{error}. The three recurrent models of LSTM are marked as LSTM-F, LSTM-P and LSTM-BI for forward standard LSTM , LSTM with peepholes and bi-directional LSTM respectively. Figure \ref{error} (d) summarizes the best error value achieved by LSTM , CNN and MLP.

From figure \ref{error}(a) and (b), the best performing models on the skiing data are the CNN with one convolution layer, the standard forward LSTM and the LSTM with peepholes. For CNN, the smaller the number of local filters used in the convolution operation, the better the learning process will be. This conclusion can be observed from the classification error value attained using only 20 filters. With few filters, the convolution operations succeeds in extracting features maps that distinguish between the two skiing gears with a classification error value of	2.4\%. By testing smaller number of filters like 10 and 15, the same error value was achieved. On the contrary, the standard forward LSTM and LSTM with peepholes have a better performance with increasing the network size (the number of LSTM units) although peepholes connections doesn't highly affect the performance especially with large number of LSTM units. %This is clear from the similar classification of both the forward LSTM and the one with peepholes.  
 By using a larger number of LSTM units at each time step, the total classification error decreased with the most minimum error of 1.6\% achieved by LSTM-F. Small number of LSTM recurrent units can also capture the dynamics of a sequence like the case of using only 25 units, however the number should be reasonable to the sequence length. Hence, Increasing recurrent units each step mostly improves the learning process because this increases the number of memory units at each step and so the number of weights to be optimized at each value of the sequence. 
 
These conclusions are also valid for the bi-directional LSTM given that only half of the total number of LSTM units are used in each of the forward and backward networks. For example, 25 units used for each of the forward and backward LSTM networks to have a total of 50 units for the bi-directional architecture. Therefore, the same behaviour is explicit again in the lower error values as utilizing more units. The worst error attained by the BLSTM, 14\%, appears with the total of 25 units, which validates the above conclusion of using too small units and its effect in declining the effectiveness. Also, the bi-directional architecture advantage of accessing future and past observations during learning didn't help the overall classification here in skiing. This can be due to the nature of skiing gears which have different intensities and frequencies , hence, a sequential learning in a forward way, like in the forward LSTM, can have better performance.

Finally, the standard multi layer perception, figure \ref{error}(c), doesn't fit the complex nature of the input skiing signals although its performance gets better with more hidden units in the hidden layer but at cost of computations. Increasing the number of hidden layers with MLP doesn’t provide a solution for better understating of the input, instead, it leads to worse performance. 

From figure \ref{error}(d) and as a conclusion over MLP, deep CNN and LSTM networks, the standard forward LSTM is the most convenient for classifying the gears of skiing data with a classification error value 1.6\%. It is the best model for this data due to many characteristics. LSTM has the advantage of shared weights between all cells which enhance the learning through the input sequence. Furthermore, it incorporates different gates that help in memorizing the dynamics of the input. All these advantages suits the temporal nature of the skiing data and hence best results are obtained.

\section*{5. Conclusions}
This paper introduced deep learning for the first time in the classification of skiing gears using sensor based signals. Different deep learning architectures were applied and compared to select the best suited with the lowest classification error. A total of five deep models were implemented, two convolutional network models and three long short term memory models, were tested in addition to two standard multi-layer perceptron (MLP) models. The final error of classification over the testing skiing data were reported using different settings of each model. 

The MLP failed to have a high performance in the classification due to the complex nature of the input skiing signals which have varying intensities and frequencies due to the different styles of skiers to perform the same gear. Both stacked layers and the non-linear complex functions used by deep models enable CNN and LSTM to achieve good results represented in low classification error. CNN had the minimum error of 2.4\% using few number of local filters for the convolution process where increasing the filters doesn't mean improving the performance. The standard forward LSTM had the lowest classification error, 1.6\%, over all the five models thanks to its recurrent architecture which utilize shared weights through all time steps. Therefore, the learned weights of memory gates at each step are propagated to the next time step till the end of each signal. Also, it can be concluded that the larger the number of LSTM units used each time step, the better the performance of the whole network but a larger computation time of training. On the other side, fewer recurrent units yields to a poor learning with a higher classification error. The best number of units here, in skiing, found to be the same as the length of the skiing input signal. 

As a future work, more experiments using other deep learning based models will be carried. Also, different datasets of skiing with a varied percentage of gears will be used to test the generality of the above conclusions. Finally, different styles within each gear should be analysed to study the different ways of performing skiing and assess the quality of performance.

%\section*{References}
\nocite{1}
\nocite{*}
\bibliographystyle{plain}
\bibliography{bibliography}

\begin{thebibliography}{10}

\bibitem{barshan2014recognizing}
Billur Barshan and Murat~Cihan Y{\"u}ksek.
\newblock Recognizing daily and sports activities in two open source machine
  learning environments using body-worn sensor units.
\newblock {\em The Computer Journal}, 57(11):1649--1667, 2014.

\bibitem{chen2015survey}
Chen Chen, Roozbeh Jafari, and Nasser Kehtarnavaz.
\newblock A survey of depth and inertial sensor fusion for human action
  recognition.
\newblock {\em Multimedia Tools and Applications}, pages 1--21, 2015.

\bibitem{chen2016lstm}
Yuwen Chen, Kunhua Zhong, Ju~Zhang, Qilong Sun, and Xueliang Zhao.
\newblock Lstm networks for mobile human activity recognition.
\newblock 2016.

\bibitem{donahue2015long}
Jeffrey Donahue, Lisa Anne~Hendricks, Sergio Guadarrama, Marcus Rohrbach,
  Subhashini Venugopalan, Kate Saenko, and Trevor Darrell.
\newblock Long-term recurrent convolutional networks for visual recognition and
  description.
\newblock In {\em Proceedings of the IEEE conference on computer vision and
  pattern recognition}, pages 2625--2634, 2015.

\bibitem{fuji2011development}
Kenichirou Fuji, Hiroki Tamura, Takaya Maeda, and Koichi Tanno.
\newblock Development of a motion analysis system using acceleration sensors
  for tennis and its evaluations.
\newblock {\em Artificial Life and Robotics}, 16(2):190--193, 2011.

\bibitem{gers2002learning}
Felix~A Gers, Nicol~N Schraudolph, and J{\"u}rgen Schmidhuber.
\newblock Learning precise timing with lstm recurrent networks.
\newblock {\em Journal of machine learning research}, 3(Aug):115--143, 2002.

\bibitem{gjoreskicomparing}
Hristijan Gjoreski, Jani Bizjak, Martin Gjoreski, and Matja{\v{z}} Gams.
\newblock Comparing deep and classical machine learning methods for human
  activity recognition using wrist accelerometer.

\bibitem{graves2005framewise}
Alex Graves and J{\"u}rgen Schmidhuber.
\newblock Framewise phoneme classification with bidirectional lstm and other
  neural network architectures.
\newblock {\em Neural Networks}, 18(5):602--610, 2005.

\bibitem{hammerla2016deep}
Nils~Y Hammerla, Shane Halloran, and Thomas Ploetz.
\newblock Deep, convolutional, and recurrent models for human activity
  recognition using wearables.
\newblock {\em arXiv preprint arXiv:1604.08880}, 2016.

\bibitem{holst2013classification}
Anders Holst and Arndt Jonasson.
\newblock Classification of movement patterns in skiing.
\newblock In {\em SCAI}, pages 115--124, 2013.

\bibitem{jiang2015human}
Wenchao Jiang and Zhaozheng Yin.
\newblock Human activity recognition using wearable sensors by deep
  convolutional neural networks.
\newblock In {\em Proceedings of the 23rd ACM international conference on
  Multimedia}, pages 1307--1310. ACM, 2015.

\bibitem{lee2012mobile}
M~Lee and S~Cho.
\newblock Mobile gesture recognition using hierarchical recurrent neural
  network with bidirectional long short-term memory.
\newblock In {\em In proceedings of UBICOMM}, pages 138--141. 2012.

\bibitem{lu2016multi}
Feng Lu, Danfeng Wang, Haoying Wu, and Wei Xie.
\newblock A multi-classifier combination method using sffs algorithm for
  recognition of 19 human activities.
\newblock In {\em International Conference on Computational Science and Its
  Applications}, pages 519--529. Springer, 2016.

\bibitem{ordonez2016deep}
Francisco~Javier Ord{\'o}{\~n}ez and Daniel Roggen.
\newblock Deep convolutional and lstm recurrent neural networks for multimodal
  wearable activity recognition.
\newblock {\em Sensors}, 16(1):115, 2016.

\bibitem{ristnercomparison}
Moa Ristner.
\newblock A comparison of a k-nearest neighbour and a markov model for
  classification of gears in cross-country skiing.

\bibitem{ronao2016human}
Charissa~Ann Ronao and Sung-Bae Cho.
\newblock Human activity recognition with smartphone sensors using deep
  learning neural networks.
\newblock {\em Expert Systems with Applications}, 59:235--244, 2016.

\bibitem{stoggl2014automatic}
Thomas St{\"o}ggl, Anders Holst, Arndt Jonasson, Erik Andersson, Tobias Wunsch,
  Christer Norstr{\"o}m, and Hans-Christer Holmberg.
\newblock Automatic classification of the sub-techniques (gears) used in
  cross-country ski skating employing a mobile phone.
\newblock {\em Sensors}, 14(11):20589--20601, 2014.

\bibitem{wang2015sport}
Wen-Fong Wang, Ching-Yu Yang, and Ji-Ting Guo.
\newblock A sport recognition method with utilizing less motion sensors.
\newblock In {\em Genetic and Evolutionary Computing}, pages 155--167.
  Springer, 2015.

\bibitem{yang2015deep}
Jian~Bo Yang, Minh~Nhut Nguyen, Phyo~Phyo San, Xiao~Li Li, and Shonali
  Krishnaswamy.
\newblock Deep convolutional neural networks on multichannel time series for
  human activity recognition.
\newblock In {\em Proceedings of the 24th International Joint Conference on
  Artificial Intelligence (IJCAI), Buenos Aires, Argentina}, pages 25--31,
  2015.

\bibitem{zeng2014convolutional}
Ming Zeng, Le~T Nguyen, Bo~Yu, Ole~J Mengshoel, Jiang Zhu, Pang Wu, and Joy
  Zhang.
\newblock Convolutional neural networks for human activity recognition using
  mobile sensors.
\newblock In {\em Mobile Computing, Applications and Services (MobiCASE), 2014
  6th International Conference on}, pages 197--205. IEEE, 2014.

\end{thebibliography}
\end{document}